\documentclass[conference]{IEEEtran}
\IEEEoverridecommandlockouts
\usepackage{cite}
\usepackage{amsmath,amssymb,amsfonts}
\usepackage{algorithmic}
\usepackage[table]{xcolor}
\usepackage{graphicx}
\usepackage{textcomp}
\usepackage{listings}
\usepackage{multirow}
\usepackage{booktabs}
\usepackage{tabularx}
\usepackage{makecell}
\usepackage{hyperref}
\usepackage{url}
\usepackage{float}
\usepackage{longtable}
\usepackage{caption}
\usepackage{xcolor}
\usepackage{colortbl}
\usepackage{booktabs} 
\def\BibTeX{{\rm B\kern-.05em{\sc i\kern-.025em b}\kern-.08em
    T\kern-.1667em\lower.7ex\hbox{E}\kern-.125emX}}
\begin{document}

\title{Architecting Clinical Collaboration: Multi-Agent Reasoning Systems for Multimodal Medical VQA\thanks{Code available at: \url{https://github.com/karishmathakrar/arc-mediqa-magic-2025/}}}



\author{\IEEEauthorblockN{Karishma Thakrar}
\IEEEauthorblockA{\textit{Georgia Institute of Technology} \\
Atlanta, GA, USA}
\and
\IEEEauthorblockN{Shreyas Basavatia}
\IEEEauthorblockA{\textit{Georgia Institute of Technology} \\
Atlanta, GA, USA}
\and
\IEEEauthorblockN{Akshay Daftardar}
\IEEEauthorblockA{\textit{Georgia Institute of Technology} \\
Atlanta, GA, USA}
}

\maketitle

\begin{abstract}

Dermatological care via telemedicine often lacks the rich context of in-person visits. Clinicians must make diagnoses based on a handful of images and brief descriptions, without the benefit of physical exams, second opinions, or reference materials. While many medical AI systems attempt to bridge these gaps with domain-specific fine-tuning, this work hypothesized that mimicking clinical reasoning processes could offer a more effective path forward. This study tested seven vision-language models on medical visual question answering across six configurations: baseline models, fine-tuned variants, and both augmented with either reasoning layers that combine multiple model perspectives, analogous to peer consultation, or retrieval-augmented generation that incorporates medical literature at inference time, serving a role similar to reference-checking. While fine-tuning degraded performance in four of seven models with an average 30\% decrease, baseline models collapsed on test data. Clinical-inspired architectures, meanwhile, achieved up to 70\% accuracy, maintaining performance on unseen data while generating explainable, literature-grounded outputs critical for clinical adoption. These findings demonstrate that medical AI succeeds by reconstructing the collaborative and evidence-based practices fundamental to clinical diagnosis.

\end{abstract}


\section{Introduction}
Fine-tuning large models on medical data, the standard approach to medical AI, assumes domain exposure produces clinical competence \cite{xu2024multi}. Yet dermatology models show 15\% performance drops in real-world settings \cite{chiou2025midas}, and catastrophic forgetting causes models to generate outputs exclusively from their training data \cite{zhai2023catastrophic}. This brittleness suggests a fundamental mismatch between current approaches and clinical reasoning. Additionally, physician groups achieve 85.6\% diagnostic accuracy versus 62.5\% for individuals \cite{sermo2024collab}, as collaboration reduces cognitive load and bias \cite{lee2024ccr}. However, logistical constraints force physicians to work alone, a problem telemedicine intensifies by eliminating physical exams, peer consultation, and immediate reference access \cite{rodriguez2024steps}. With consultation volumes surging \cite{klas2025drowning}, AI that restores collaborative capabilities becomes critical.

This paper presents AI systems that structurally implement clinical reasoning rather than merely training on medical data. The proposed architectures incorporate three expert capabilities: domain knowledge, collaborative reasoning through multiple perspectives, and dynamic literature verification. Seven vision-language models were evaluated across six configurations: baseline, fine-tuned, and both enhanced with either multi-model reasoning layers (emulating peer consultation) or agentic RAG (emulating reference-checking). On a noisy dermatological VQA dataset, fine-tuning degraded four of seven models while some baseline models collapsed from 67\% to 37\% accuracy on test data. Clinical-inspired architectures maintained \>70\% accuracy across datasets while generating explainable, literature-grounded outputs, providing the transparency essential for clinical trust in minimally supervised telemedicine contexts \cite{doi:10.1056/AIe2401235, primiero2024tbp}. These findings suggest architectural designs that emulate collaborative clinical practices offer a robust alternative to fine-tuning, providing practical solutions for clinically constrained telemedicine environments.

\section{Related Work}

Medical vision-language models follow two paradigms: domain-specific fine-tuning versus architectural approaches. Fine-tuning methods like Med-Gemini \cite{yang2024medgemini}, SkinGPT-4 \cite{zhou2023skingpt4}, and MedCoT \cite{liu2024medcot} achieve strong performance but suffer from high compute costs, limiting their practicality, or catastrophic forgetting \cite{zhai2023catastrophic}, with some real-world dermatology deployments suffering performance drops as a result. Architectural innovations offer robust alternatives; AMIE \cite{amie2024} matched physician performance through structured reasoning, though its text-only design limits multimodal applications beyond primary care. These successes suggest that emulating clinical processes architecturally may be superior to forcing domain knowledge through fine-tuning.

Recent advances reinforce the architectural paradigm. Chain-of-thought prompting improved medical VQA performance by 16-19\% across multiple vision-language models \cite{faira2025analyzing}, while enhancing diagnostic transparency \cite{savage2024diagnostic}. RAG systems gained 18\% over chain-of-thought through knowledge augmentation \cite{xiong2024mirage}. Yet these reamin fragmented: reasoning systems  often lack external knowledge, while traditional RAG typically employs fixed retrievals instead of dynamic, reasoning-guided information gathering \cite{sohn2024ragrag}. Multi-image reasoning presents additional challenges \cite{yu2025medframeqa}, with most systems failing to aggregate evidence across images as clinicians do. Notably, zero-shot approaches with proper reasoning architectures can match fine-tuned models in specific tasks like cancer detection \cite{sushil2024comparative}, suggesting architectural design may matter more than domain-specific training. While individual components show promise, no systematic comparison exists between clinical workflow emulation and fine-tuning across diverse VLM architectures. 

\section{Exploratory Data Analysis}
The DermaVQA-DAS dataset, adapted for the ImageCLEF2025 challenge \cite{DermaVQA-DAS} and used in this study consists of 300 unique patient encounters representing complete dermatological cases. Each encounter includes patient-level context (query titles and clinical descriptions), multiple dermatological images (approximately three per case), and structured diagnostic annotations spanning 27 questions grouped into 9 clinical question families. These cover assessment domains including body coverage extent, anatomical location, lesion morphology, and attributes like color, quantity, and texture. DermaVQA captures authentic telemedicine data with its inherent complexities; of the 2,700 total encounter-question combinations, only 6.56\% had multiple valid answers, primarily in anatomical location (31\% of encounters) and morphology questions (25\%), reflecting clinical scenarios where multiple descriptors accurately apply.

\begin{figure}[t] 
\centering 
\includegraphics[width=0.45\textwidth]{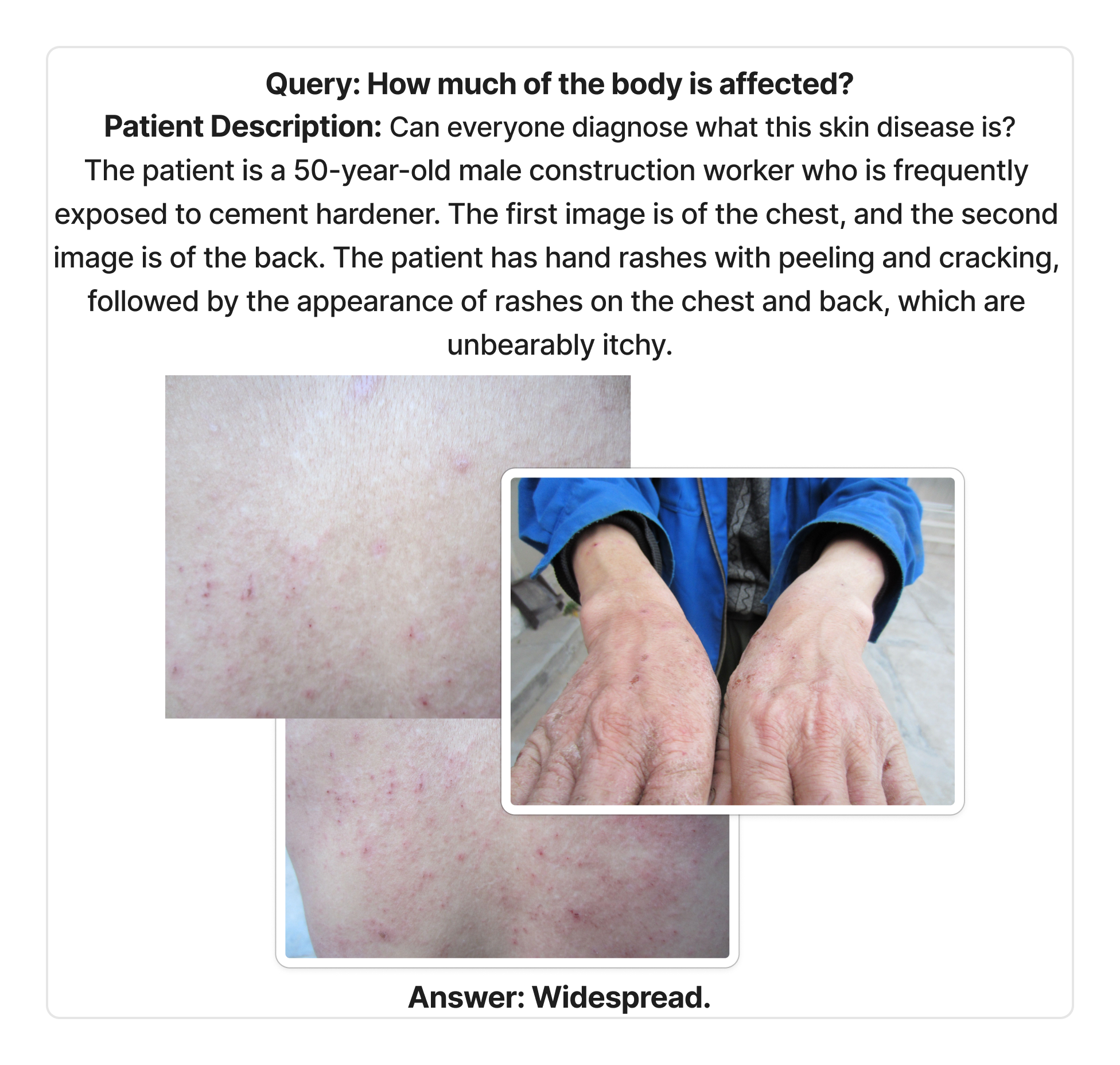} 
\caption{Example from DermaVQA-DAS dataset (ENC00006, CQID010) showing a dermatological case with multiple images of widespread rashes on chest, back, and hands.}
\label{fig:clinical-case} 
\end{figure}

The dataset authentically represents telemedicine challenges. Patient contexts are user-generated with varying detail levels and informal language, mirroring real patient-submitted information (see Figure~\ref{fig:clinical-case}). Image quality ranges from sharp clinical photos to blurry user submissions, typical of remote consultations. Questions include closely related answer choices, such as "red" versus "pink" lesions or "raised" versus "bumpy" morphologies, that reflect the subtle distinctions clinicians must make in practice. These complexities were addressed through a generative approach using LLMs with augmented reasoning and agentic RAG, enabling robust handling of the clinical ambiguity and variable input quality typical of telemedicine encounters.

\section{Methodology}

Six architectural configurations for medical visual question answering were evaluated, systematically comparing baseline and fine-tuned vision-language models both independently and as components within enhanced reasoning frameworks. Each configuration was tested across seven open-source models: LLaMA-3.2-11B-Vision, Qwen2-VL (2B, 7B), Qwen2.5-VL (3B, 7B), and Gemma-3 (4B, 12B). For agentic reasoning and retrieval-augmented generation (RAG), Gemini 2.5 Flash was additionally employed as the instruction-following model responsible for multi-step reasoning and aggregation.

\begin{enumerate}
\item Baseline model inference
\item Fine-tuned models (LoRA-adapted) inference
\item Reasoning layer using baseline model predictions
\item Reasoning layer using fine-tuned model predictions
\item Agentic RAG system using baseline model predictions
\item Agentic RAG system using fine-tuned model predictions
\end{enumerate}

Configurations 1-2 perform direct inference, while 3-6 aggregate predictions from all seven models with agentic RAG additionally incorporating medical literature retrieval.

\subsection{Model Fine-tuning \& Inference}

Seven vision-language models were fine-tuned on DermaVQA-DAS: LLaMA-3.2-11B-Vision-Instruct, Gemma-3 (4B/12B), and Qwen2/2.5-VL (2B/3B/7B variants). Training employed parameter-efficient fine-tuning using LoRA (rank 8, alpha 16, dropout 0.05) with 4-bit NF4 quantization to optimize memory usage. Models were trained for 3 epochs using AdamW optimizer (learning rate 1e-4), with gradient accumulation achieving an effective batch size of 32. The training objective minimized the standard autoregressive language modeling loss:
$$\mathcal{L} = -\frac{1}{N} \sum_{i=1}^{N} \sum_{t=1}^{T_i} \log p(y_{i,t} | y_{i,<t}, x_i)$$
where $N$ is the number of samples, $T_i$ is the length of the target answer for sample $i$, $y_{i,t}$ is the $t$-th token in the answer, and $x_i$ is the input (image and prompt). Special tokens and padding were masked with -100 to exclude them from loss computation. For multi-label questions, the loss was computed over the comma-separated sequence of answers, with the model learning to generate all valid answers as a single text output. Training on NVIDIA A100 80GB GPUs required approximately 10 hours per model (70 GPU-hours total). The \texttt{SFTTrainer} from \texttt{TRL} library managed the training process with checkpoints saved every 50 steps. After training, LoRA adapters were merged into base models to create standalone versions for inference.

The inference pipeline evaluated both baseline and fine-tuned models, loading all models with 4-bit NF4 quantization to ensure consistent memory usage and fair comparison across model configurations. For each test sample, the system constructed prompts combining clinical context, question text, and answer options with the corresponding medical image. Generation parameters were set to temperature 0.9, top-p 0.95, and top-k 64. Since each image was processed independently, predictions from multiple images of the same encounter were aggregated using a frequency-based mechanism that respected question-specific maximum answer limits, with the most common responses selected up to the allowed limit.

\subsection{Reasoning Layer}
The reasoning layer operates as a standalone enhancement architecture that receives predictions from all seven vision-language models (either baseline or fine-tuned versions) for each encounter-question pair, along with the original images and clinical context. It emulates a senior dermatologist synthesizing multiple expert opinions, performing interpretive synthesis rather than simple majority voting. Using \texttt{gemini-2.5-flash-preview} \cite{gemini2.5flash}, the reasoning layer utilizes a three-stage clinical reasoning process:

\textbf{Stage 1: Image Analysis.} The model extracts standardized features across ten clinical dimensions (size, location, morphology, color, count, extent, texture, onset indicators, itch indicators, and overall impression) from each dermatological image. When multiple images exist per encounter, individual analyses are aggregated to identify consistent patterns across images while retaining image-specific findings.

\textbf{Stage 2: Context Extraction.} Patient-provided clinical descriptions from the dataset are parsed into thirteen structured categories including demographics, symptom progression, triggers, and medical history, enabling systematic integration with visual findings.

\textbf{Stage 3: Evidence Synthesis.} The model integrates analyses from prior stages with predictions from all seven vision-language models. Model predictions are treated as advisory inputs with explicit instructions to maintain critical evaluation rather than assuming consensus equals correctness. Query-specific prompts enforce appropriate answer constraints: multiple selections for multi-label questions, single selection for exclusive-choice questions.

\subsection{Agentic Retrieval-Augmented Generation}
A separate multi-agent retrieval-augmented generation system was developed using Gemini 2.5 Flash \cite{gemini2.5flash} that emulates how dermatologists combine visual assessment with targeted reference consultation. The system distributes tasks across specialized agents that adapt behavior based on evidence availability, diagnostic ambiguity, and question complexity.

\begin{figure*}[t]
\centering
\includegraphics[width=1.0\textwidth]{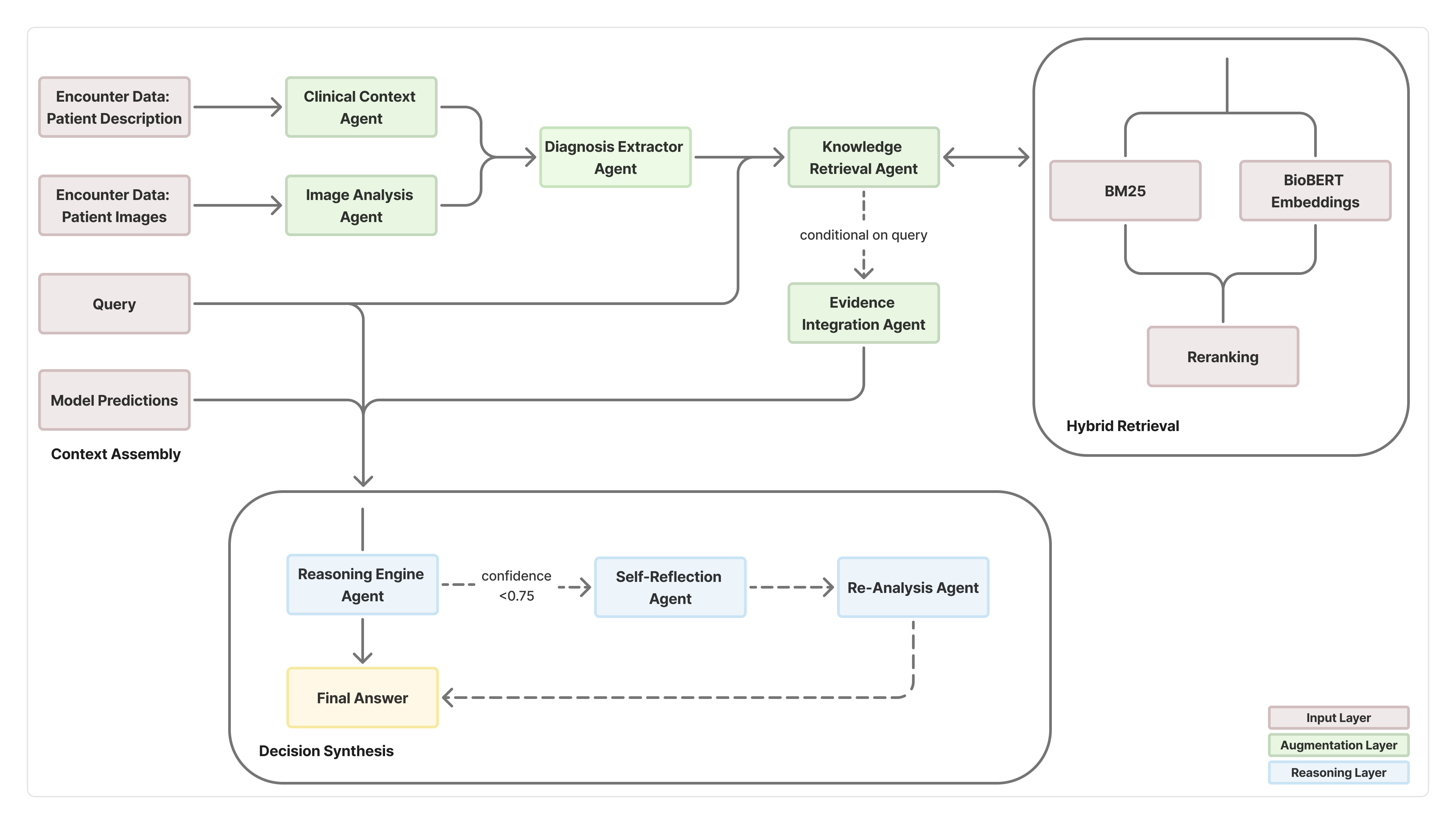}
\caption{Agentic RAG architecture featuring a multi-stage pipeline with agents that process queries, encounter data, and model predictions through collaborative reasoning and iterative refinement.}
\label{fig:ar-architecture-label}
\end{figure*}

\subsubsection{Input Layer}
The input layer collects patient images, symptom descriptions, diagnostic predictions from all seven vision-language models (baseline or fine-tuned versions), and a curated medical knowledge base containing 800+ dermatology entries from the American Academy of Dermatology \cite{brucewayne0459_skin_2024}.

\subsubsection{Context Assembly}
Specialized agents operate in sequence to enable adaptive multi-stage decision-making:

\textbf{Image Analysis Agent} extracts structured features (size, location, shape, color, distribution) from dermatological images, identifying visual indicators of duration and trauma.

\textbf{Clinical Context Agent} processes patient-authored text into structured medical categories (demographics, symptoms, duration, triggers), filtering noise while preserving clinically relevant information.

\textbf{Diagnosis Extractor and Knowledge Retrieval Agents} synthesize outputs from previous agents to generate diagnostic hypotheses. These hypotheses guide targeted search queries specific to both the suspected condition and question type, ensuring retrieved knowledge directly addresses the clinical query. The system employs hybrid retrieval combining BM25 keyword matching with semantic search using clinical text embeddings \cite{deka2022evidence}. Retrieved passages are reranked using a cross-encoder \cite{crossencoder_msmarco_minilm_l6_v2} to prioritize the most relevant information. For purely visual questions (color, size), retrieval is disabled as direct image analysis suffices.

\textbf{Evidence Integration Agent} synthesizes outputs from all prior agents using adaptive task-specific weights, emphasizing visual cues for appearance questions, clinical history for treatment queries, and literature for diagnosis tasks.

\subsubsection{Decision Synthesis}
\textbf{Reasoning Engine Agent} combines integrated evidence with model predictions to generate answers, confidence scores (0-1), and structured explanations.

\textbf{Self-Reflection Agent} evaluates outputs when confidence falls below a specified threshold, identifying potential reasoning gaps or overlooked evidence.

\textbf{Re-Analysis Agent} activates only when self-reflection identifies issues requiring revision, conducting deeper reassessment to generate improved answers.

This architecture enables autonomous error correction through iterative clinical reasoning, making predictions, evaluating confidence, and self-correcting when needed. Both clinical-inspired architectures operate through structured inference-time reasoning without additional training, prioritizing explainability and adaptability for deployment in clinical settings.

\subsection{Evaluation Metrics}
Performance was evaluated using a partial-credit metric for multi-label questions: $\text{accuracy} = \frac{|P \cap G|}{\max(|P|, |G|)}$, where $P$ and $G$ represent predicted and ground truth answer sets. This Jaccard-based approach awards partial credit for overlapping predictions rather than requiring exact matches, with overall accuracy averaged across all question-encounter pairs. This reflects clinical reality where multiple descriptors often apply to dermatological presentations.

\section{Results}

The evaluation revealed a major challenge in model deployment: distribution shift between validation and test datasets (\autoref{tab:base-model-comparison}). While Qwen2.5-VL-7B achieved 67.21\% validation accuracy, its test performance dropped to 37.43\%, a 29.78\% decrease. In contrast, the Reasoning Layer maintained performance across datasets (71.23\% validation, 70.62\% test), as did Agentic RAG (68.98\% validation, 69.24\% test). Both architectures nearly doubled the baseline model's test performance. Preliminary experiments with Gemini 2.5 Flash alone also revealed issues with direct inference, producing contradictory outputs such as selecting both ``smooth" and ``rough" for texture (ENC00858, CQID036) or both ``yes" and ``no" for itching (ENC00868, CQID025), further motivating the need for structured reasoning approaches.

Baseline model performance on validation data varied substantially (\autoref{tab:base-models}). Qwen2.5-VL-7B led at 67.21\%, followed by Qwen2.5-VL-3B (55.56\%) and Qwen2-VL-7B (54.75\%). Model size showed weak correlation with accuracy: Gemma3-12B achieved 35.27\% despite having the most parameters, while smaller models like Gemma3-4B reached 54.57\%. Fine-tuning produced mixed results (\autoref{tab:finetuned-models}), improving only three models: Gemma3-12B (+10.87\%), Qwen2-VL-2B (+3.83), and Qwen2.5-VL-3B (+2.40), while degrading the others. The top-performing baseline model, Qwen2.5-VL-7B, dropped to 38.12\% after fine-tuning. When fine-tuned model predictions served as inputs to the architectures (\autoref{tab:finetuned-model-comparison}), performance decreased slightly compared to using baseline models, indicating that while architectural synthesis remained fairly robust, noise from degraded fine-tuned models can propagate through the system.

Question-level analysis revealed performance patterns based on answer type. Both augmented architectures achieved high test accuracy on questions with objective criteria: itching presence (CQID025, 87.0\% Reasoning Layer, 84.0\% Agentic RAG), anatomical location (CQID011, 84.0\% Reasoning Layer, 85.5\% Agentic RAG), and onset timing (CQID015, 85.0\% both architectures). Questions requiring subjective visual assessment showed lower performance: body coverage extent (CQID010, 51.0\% Reasoning Layer, 47.0\% Agentic RAG), lesion morphology (CQID020, 55.9\% Reasoning Layer, 55.6\% Agentic RAG), and color classification (CQID034, 55.0\% Reasoning Layer, 51.0\% Agentic RAG). This pattern held regardless of whether questions allowed single or multiple answers. Overall, the architectures showed comparable performance, with differences typically ranging 1 to 4\% across questions.

\begin{table*}[t]
\centering
\caption{Performance comparison across three approaches by question (QID): baseline model inference, Reasoning Layer, and Agentic RAG (both using baseline predictions).}
\label{tab:base-model-comparison}
\renewcommand{\arraystretch}{1.3}
\begin{tabular}{l>{\centering\arraybackslash}p{1.2cm}>{\centering\arraybackslash}p{1.2cm}>{\centering\arraybackslash}p{1.2cm}>{\centering\arraybackslash}p{1.2cm}>{\centering\arraybackslash}p{1.2cm}>{\centering\arraybackslash}p{1.2cm}}
\toprule
\multirow{2}{*}{\textbf{QID}} & \multicolumn{3}{c}{\textbf{Validation}} & \multicolumn{3}{c}{\textbf{Test}} \\
\cmidrule(lr){2-4} \cmidrule(lr){5-7}
& \textbf{Qwen2.5-VL-7B} & \textbf{Reasoning Layer} & \textbf{Agentic RAG} & \textbf{Qwen2.5-VL-7B} & \textbf{Reasoning Layer} & \textbf{Agentic RAG} \\
\midrule
CQID010 & 0.4821 & \textbf{0.5714} & 0.5357 & 0.3100 & \textbf{0.5100} & 0.4700 \\
CQID011 & 0.8333 & \textbf{0.9048} & 0.8762 & 0.3847 & 0.8403 & \textbf{0.8552} \\
CQID012 & 0.6086 & \textbf{0.7083} & 0.7009 & 0.5317 & \textbf{0.6967} & 0.6900 \\
CQID015 & 0.7679 & \textbf{0.8929} & 0.8571 & 0.3100 & \textbf{0.8500} & \textbf{0.8500} \\
CQID020 & \textbf{0.5708} & 0.5653 & 0.5771 & 0.3122 & \textbf{0.5587} & 0.5561 \\
CQID025 & \textbf{0.8929} & 0.8036 & 0.8036 & 0.4200 & \textbf{0.8700} & 0.8400 \\
CQID034 & \textbf{0.4643} & 0.4286 & 0.3929 & 0.0100 & \textbf{0.5500} & 0.5100 \\
CQID035 & 0.8750 & \textbf{0.8929} & 0.8214 & 0.7200 & 0.8100 & \textbf{0.8200} \\
CQID036 & 0.5536 & \textbf{0.6429} & \textbf{0.6429} & 0.3700 & \textbf{0.6700} & 0.6400 \\
\midrule
\textbf{Average} & 0.6721 & \textbf{0.7123} & 0.6898 & 0.3743 & \textbf{0.7062} & 0.6924 \\
\bottomrule
\end{tabular}
\end{table*}

\begin{table*}[t]
\centering
\caption{Performance of Reasoning Layer and Agentic RAG using fine-tuned model predictions by question (QID). Fine-tuned Qwen2.5-VL-7B was excluded from test inference due to overfitting (see \autoref{tab:finetuned-models}).}
\label{tab:finetuned-model-comparison}
\renewcommand{\arraystretch}{1.3}
\begin{tabular}{l>{\centering\arraybackslash}p{1.8cm}>{\centering\arraybackslash}p{1.8cm}>{\centering\arraybackslash}p{1.8cm}>{\centering\arraybackslash}p{1.8cm}}
\toprule
\multirow{2}{*}{\textbf{QID}} & \multicolumn{2}{c}{\textbf{Validation}} & \multicolumn{2}{c}{\textbf{Test}} \\
\cmidrule(lr){2-3} \cmidrule(lr){4-5}
& \textbf{Reasoning Layer} & \textbf{Agentic RAG} & \textbf{Reasoning Layer} & \textbf{Agentic RAG} \\
\midrule
CQID010 & \textbf{0.6071} & 0.5536 & \textbf{0.5300} & 0.4400 \\
CQID011 & 0.8777 & \textbf{0.8795} & \textbf{0.8683} & 0.8363 \\
CQID012 & 0.6815 & \textbf{0.7173} & 0.6625 & \textbf{0.6858} \\
CQID015 & \textbf{0.8214} & \textbf{0.8214} & \textbf{0.8100} & 0.7800 \\
CQID020 & \textbf{0.5821} & 0.5421 & \textbf{0.5649} & 0.5544 \\
CQID025 & \textbf{0.8214} & 0.8036 & \textbf{0.8900} & 0.8600 \\
CQID034 & \textbf{0.4643} & 0.3929 & \textbf{0.6000} & 0.4800 \\
CQID035 & \textbf{0.8929} & 0.8393 & \textbf{0.8100} & 0.7900 \\
CQID036 & \textbf{0.6071} & \textbf{0.6071} & \textbf{0.6500} & \textbf{0.6500} \\
\midrule
\textbf{Average} & \textbf{0.7062} & 0.6841 & \textbf{0.7095} & 0.6752 \\
\bottomrule
\end{tabular}
\end{table*}

\subsection{Ablation Studies}
Ablation studies were conducted to explore the contribution of individual components. Results demonstrated that the full architectural design works synergistically, with performance degrading when using partial model ensembles (2-4 models: 58.9 - 59.9\%) or removing self-reflection (59.8\%). Confidence threshold experiments (0.6 - 0.9) showed lower performance (58.6 - 62.2\%) compared to the chosen threshold of 0.75, which achieved 69.2\% accuracy, validating the architectural choices.

\definecolor{muted-green}{RGB}{220, 240, 220}
\definecolor{dark-green}{RGB}{40, 120, 40}
\definecolor{muted-red}{RGB}{255, 230, 230}
\definecolor{dark-red}{RGB}{180, 60, 60}

\begin{table*}[t]
\centering
\caption{Baseline model performance by question (QID) across seven vision-language models on validation data. Qwen2.5-VL-7B achieves the highest accuracy.}
\label{tab:base-models}
\renewcommand{\arraystretch}{1.3}
\begin{tabular}{l>{\centering\arraybackslash}p{1.1cm}>{\centering\arraybackslash}p{1.1cm}>{\centering\arraybackslash}p{1.1cm}>{\centering\arraybackslash}p{1.1cm}>{\centering\arraybackslash}p{1.1cm}>{\centering\arraybackslash}p{1.1cm}>{\centering\arraybackslash}p{1.3cm}>{\centering\arraybackslash}p{1.3cm}}
\toprule
\textbf{QID} & \textbf{Qwen2-VL-2B} & \textbf{Qwen2-VL-7B} & \textbf{Qwen2.5-VL-3B} & \textbf{Qwen2.5-VL-7B} & \textbf{Gemma3-4B} & \textbf{Gemma3-12B} & \textbf{LLaMA-3.2-11B} & \textbf{Question Avg.} \\
\midrule
CQID010 & 0.2115 & 0.4107 & \textbf{0.6250} & 0.4821 & 0.5179 & 0.1071 & 0.3036 & 0.3797 \\
CQID011 & 0.6869 & 0.7863 & 0.7417 & \textbf{0.8333} & 0.8256 & 0.7045 & 0.6821 & 0.7515 \\
CQID012 & 0.5737 & 0.6250 & \textbf{0.6696} & 0.6086 & 0.6280 & 0.5179 & 0.6161 & 0.6056 \\
CQID015 & 0.2500 & 0.4821 & 0.5000 & \textbf{0.7679} & 0.5714 & 0.3750 & 0.3393 & 0.4694 \\
CQID020 & 0.4641 & 0.5698 & 0.5002 & 0.5708 & \textbf{0.6188} & 0.5589 & 0.4051 & 0.5268 \\
CQID025 & 0.4038 & 0.7500 & 0.5357 & \textbf{0.8929} & 0.6250 & 0.0714 & 0.4643 & 0.5347 \\
CQID034 & 0.3269 & 0.3571 & 0.4286 & \textbf{0.4640} & 0.4643 & 0.1964 & 0.4286 & 0.3808 \\
CQID035 & 0.0577 & 0.5714 & 0.6786 & \textbf{0.8750} & 0.2500 & 0.3393 & 0.6607 & 0.4904 \\
CQID036 & 0.3077 & 0.3750 & 0.3214 & \textbf{0.5536} & 0.4107 & 0.3036 & 0.4107 & 0.3832 \\
\midrule
\textbf{Avg.} & 0.3647 & 0.5475 & 0.5556 & \textbf{0.6721} & 0.5457 & 0.3527 & 0.4790 & 0.5025 \\
\bottomrule
\end{tabular}
\end{table*}

\begin{table*}[t]
\centering
\caption{Fine-tuned model performance by question (QID) on validation dataset. Green cells indicate improvement from baseline; red indicates degradation. Qwen2.5-VL-3B achieves highest accuracy.}
\label{tab:finetuned-models}
\renewcommand{\arraystretch}{1.3}
\begin{tabular}{l>{\centering\arraybackslash}p{1.1cm}>{\centering\arraybackslash}p{1.1cm}>{\centering\arraybackslash}p{1.1cm}>{\centering\arraybackslash}p{1.1cm}>{\centering\arraybackslash}p{1.1cm}>{\centering\arraybackslash}p{1.1cm}>{\centering\arraybackslash}p{1.3cm}>{\centering\arraybackslash}p{1.3cm}}
\toprule
\textbf{QID} & \textbf{Qwen2-VL-2B} & \textbf{Qwen2-VL-7B} & \textbf{Qwen2.5-VL-3B} & \textbf{Qwen2.5-VL-7B} & \textbf{Gemma3-4B} & \textbf{Gemma3-12B} & \textbf{LLaMA-3.2-11B} & \textbf{Question Avg.} \\
\midrule
CQID010 & \cellcolor{muted-green}\textcolor{dark-green}{0.3600} & \cellcolor{muted-red}\textcolor{dark-red}{0.2321} & \cellcolor{muted-green}\textcolor{dark-green}{\textbf{0.6607}} & \cellcolor{muted-red}\textcolor{dark-red}{0.2500} & \cellcolor{muted-green}\textcolor{dark-green}{0.5714} & \cellcolor{muted-green}\textcolor{dark-green}{0.2500} & \cellcolor{muted-green}\textcolor{dark-green}{0.3571} & \cellcolor{muted-green}\textcolor{dark-green}{0.3830} \\
CQID011 & \cellcolor{muted-red}\textcolor{dark-red}{0.6747} & \cellcolor{muted-red}\textcolor{dark-red}{0.3667} & \cellcolor{muted-green}\textcolor{dark-green}{0.7634} & \cellcolor{muted-red}\textcolor{dark-red}{0.3839} & \cellcolor{muted-green}\textcolor{dark-green}{0.8360} & \cellcolor{muted-green}\textcolor{dark-green}{\textbf{0.8435}} & \cellcolor{muted-red}\textcolor{dark-red}{0.4976} & \cellcolor{muted-red}\textcolor{dark-red}{0.6237} \\
CQID012 & \cellcolor{muted-red}\textcolor{dark-red}{0.5567} & \cellcolor{muted-red}\textcolor{dark-red}{0.5119} & \cellcolor{muted-green}\textcolor{dark-green}{\textbf{0.6726}} & \cellcolor{muted-red}\textcolor{dark-red}{0.5119} & \cellcolor{muted-green}\textcolor{dark-green}{0.6577} & \cellcolor{muted-green}\textcolor{dark-green}{0.5417} & \cellcolor{muted-red}\textcolor{dark-red}{0.5506} & \cellcolor{muted-red}\textcolor{dark-red}{0.5719} \\
CQID015 & \cellcolor{muted-green}\textcolor{dark-green}{0.3200} & \cellcolor{muted-red}\textcolor{dark-red}{0.0893} & \cellcolor{muted-green}\textcolor{dark-green}{0.5357} & \cellcolor{muted-red}\textcolor{dark-red}{0.2857} & \cellcolor{muted-red}\textcolor{dark-red}{0.4107} & \cellcolor{muted-green}\textcolor{dark-green}{\textbf{0.8036}} & \cellcolor{muted-red}\textcolor{dark-red}{0.2857} & \cellcolor{muted-red}\textcolor{dark-red}{0.3901} \\
CQID020 & \cellcolor{muted-red}\textcolor{dark-red}{0.4560} & \cellcolor{muted-red}\textcolor{dark-red}{0.2942} & \cellcolor{muted-green}\textcolor{dark-green}{0.5126} & \cellcolor{muted-red}\textcolor{dark-red}{0.3031} & \cellcolor{muted-green}\textcolor{dark-green}{\textbf{0.6253}} & \cellcolor{muted-green}\textcolor{dark-green}{0.6071} & \cellcolor{muted-red}\textcolor{dark-red}{0.2990} & \cellcolor{muted-red}\textcolor{dark-red}{0.4425} \\
CQID025 & \cellcolor{muted-green}\textcolor{dark-green}{0.4600} & \cellcolor{muted-red}\textcolor{dark-red}{\textbf{0.6250}} & \cellcolor{muted-red}\textcolor{dark-red}{0.4821} & \cellcolor{muted-red}\textcolor{dark-red}{0.3929} & \cellcolor{muted-red}\textcolor{dark-red}{0.2500} & \cellcolor{muted-green}\textcolor{dark-green}{0.0893} & \cellcolor{muted-red}\textcolor{dark-red}{0.4107} & \cellcolor{muted-red}\textcolor{dark-red}{0.3871} \\
CQID034 & \cellcolor{muted-red}\textcolor{dark-red}{0.2600} & \cellcolor{muted-red}\textcolor{dark-red}{0.1250} & \cellcolor{muted-red}\textcolor{dark-red}{0.3571} & \cellcolor{muted-red}\textcolor{dark-red}{0.0357} & \cellcolor{muted-red}\textcolor{dark-red}{0.3036} & \cellcolor{muted-green}\textcolor{dark-green}{0.3214} & \cellcolor{muted-red}\textcolor{dark-red}{\textbf{0.3750}} & \cellcolor{muted-red}\textcolor{dark-red}{0.2540} \\
CQID035 & \cellcolor{muted-green}\textcolor{dark-green}{0.3200} & \cellcolor{muted-red}\textcolor{dark-red}{0.1250} & \cellcolor{muted-green}\textcolor{dark-green}{\textbf{0.8214}} & \cellcolor{muted-red}\textcolor{dark-red}{0.7679} & \cellcolor{muted-green}\textcolor{dark-green}{0.4821} & \cellcolor{muted-green}\textcolor{dark-green}{0.5179} & \cellcolor{muted-red}\textcolor{dark-red}{0.5714} & \cellcolor{muted-green}\textcolor{dark-green}{0.5151} \\
CQID036 & \cellcolor{muted-red}\textcolor{dark-red}{0.2200} & \cellcolor{muted-red}\textcolor{dark-red}{0.2321} & \cellcolor{muted-green}\textcolor{dark-green}{0.4107} & \cellcolor{muted-red}\textcolor{dark-red}{\textbf{0.5000}} & \cellcolor{muted-red}\textcolor{dark-red}{0.0714} & \cellcolor{muted-red}\textcolor{dark-red}{0.1786} & \cellcolor{muted-red}\textcolor{dark-red}{0.2500} & \cellcolor{muted-red}\textcolor{dark-red}{0.2661} \\
\midrule
\textbf{Avg.} & \cellcolor{muted-green}\textcolor{dark-green}{0.4030} & \cellcolor{muted-red}\textcolor{dark-red}{0.2890} & \cellcolor{muted-green}\textcolor{dark-green}{\textbf{0.5796}} & \cellcolor{muted-red}\textcolor{dark-red}{0.3812} & \cellcolor{muted-red}\textcolor{dark-red}{0.4676} & \cellcolor{muted-green}\textcolor{dark-green}{0.4615} & \cellcolor{muted-red}\textcolor{dark-red}{0.3997} & \cellcolor{muted-red}\textcolor{dark-red}{0.4259} \\
\bottomrule
\end{tabular}
\end{table*}

\section{Discussion}
The clinical-inspired architectures maintained up to 70\% accuracy across distribution shifts that degraded single model performance by 30\%, demonstrating that structural approaches can be more robust than parameter adaptation. The Reasoning Layer achieved 70.62\% test accuracy compared to 37.43\% for the best baseline model, while Agentic RAG reached 69.24\%, within 2\% of the Reasoning Layer. This robustness was achieved without fine-tuning, using only pre-trained models as components within structured reasoning frameworks.

\begin{figure}[t]
    \centering
    \includegraphics[width=1.0\linewidth]{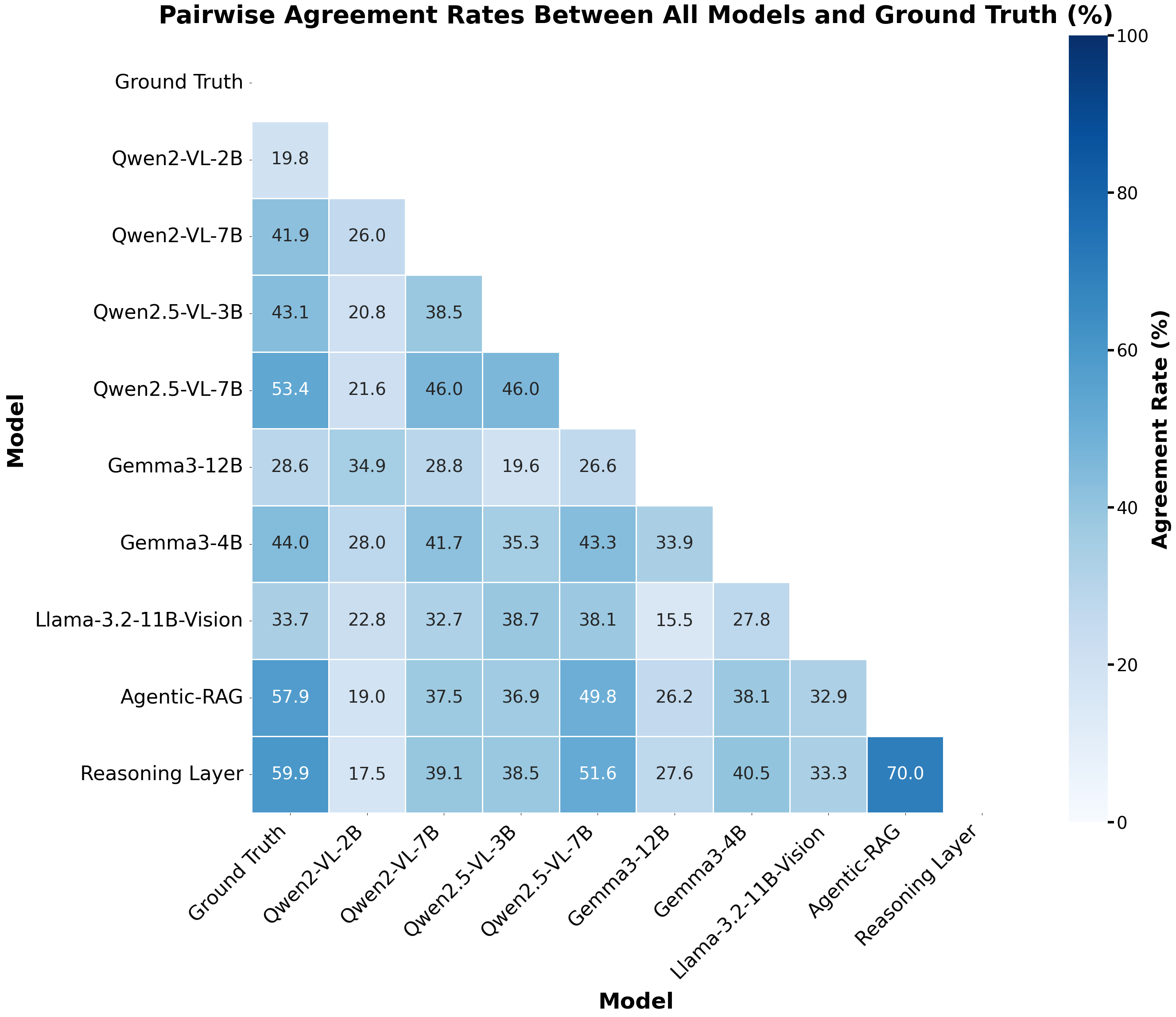}
    \caption{Pairwise agreement rates (\%) among baseline model predictions, augmented architectures, and ground truth labels. Higher values indicate greater consistency in labels.}
    \label{fig:output1}
\end{figure}

The effectiveness of these architectures stems from successful aggregation of diverse model predictions. Analysis of pairwise agreement between models revealed agreement rates ranging from 17.5\% to 51.6\% (\autoref{fig:output1}), indicating sufficient diversity for ensemble benefits where uncorrelated errors cancel rather than compound. This finding aligns with clinical literature showing physician groups achieve 85.6\% diagnostic accuracy compared to 62.5\% for individuals \cite{sermo2024collab}. The Reasoning Layer leverages this diversity by synthesizing multiple perspectives; it differs from traditional ensembles by implementing structured clinical reasoning across three stages: image analysis, context extraction, and evidence synthesis (\autoref{tab:rl-ex}). Rather than treating model outputs as votes, it frames them as advisory inputs requiring critical evaluation. This approach proved particularly effective for questions where visual and textual evidence must be integrated, such as determining lesion extent across multiple body sites.

The Agentic RAG architecture extends beyond static retrieval systems through diagnosis-driven query generation and iterative refinement. By extracting potential diagnoses from visual analysis, the system generates targeted queries that retrieve relevant medical knowledge. The self-reflection mechanism, triggered when confidence falls below 0.75, enables the system to identify and correct reasoning errors iteratively and autonomously, mirroring how physicians reconsider diagnoses when uncertain. As shown in \autoref{tab:agr-ex-3}, this produces explanations grounded in medical literature, addressing the ``black box" criticism that limits clinical AI adoption. While retrieval didn't always improve answer accuracy, it consistently enhanced answer quality by providing explainable, literature-supported rationales that allow clinicians to evaluate the reasoning process.

These approaches offer practical alternatives for institutions lacking resources for extensive fine-tuning. While the architectures require more inference computation than single models, they avoid the training costs and brittleness of fine-tuning. The trade-off favors scenarios where model robustness and explainability outweigh inference speed, particularly in telemedicine settings where accuracy and trust are paramount. Additionally, both architectures can incorporate new medical knowledge without retraining, simply by updating the knowledge base or adding new models to the ensemble. 


The architectures address key constraints in telemedicine by computationally restoring collaborative diagnosis and reference access. For isolated physicians making decisions from limited images, the Reasoning Layer provides the benefits of peer consultation while agentic RAG enables evidence-based deliberative verification. The explainable outputs allow clinicians to evaluate the reasoning process rather than trusting opaque predictions. These systems, most optimally, complement, and not replace, clinical judgment, serving as decision support tools that enhance rather than automate diagnosis.

\section{Future Work}
While the system demonstrated strong performance, several areas merit further investigation. To address inference efficiency constraints, future work should explore specific optimizations such as model distillation, selective retrieval for uncertain cases, and parallel processing architectures. Beyond technical improvements, clinical validation with practicing dermatologists is essential to assess utility and trustworthiness in actual diagnostic workflows. Testing on diverse external datasets would further establish generalizability, particularly given the findings about distribution shift vulnerability. The architectural principles demonstrated, collaborative reasoning and evidence-based decision-making, could extend to other visual medical domains such as radiology or pathology, requiring domain-specific knowledge bases and specialized feature extraction modules. Furthermore, future evaluation frameworks should incorporate measures beyond accuracy: explanation quality, reasoning coherence, and practical value to clinicians. Future evaluation should assess not only whether AI selects correct diagnoses but also whether it provides clear reasoning that helps physicians make informed decisions, essential requirements for clinical adoption in telemedicine.

\section{Conclusion}
This work demonstrates that architectural innovations encoding clinical reasoning can outperform traditional fine-tuning approaches in medical vision-language tasks. The implementation of structured reasoning layers and agentic RAG systems achieved robust performance while avoiding the instability and computational costs of domain-specific fine-tuning. The approach directly addresses the critical barrier of AI opacity in healthcare by providing traceable, literature-grounded explanations that transform AI from black-box predictor to explainable clinical tool.

The work addresses the growing need for robust AI support in telemedicine settings. The framework, combining collaborative reasoning with dynamic knowledge retrieval, offers a practical path toward trustworthy medical AI that complements rather than replaces physician judgment. These architectural principles, structured reasoning and evidence-based explanation, provide a foundation for developing transparent AI systems in clinical settings where explainability is paramount.

\onecolumn
\appendix

\section{Appendix}

\begin{figure}[H]
    \centering
    \includegraphics[width=0.42\textwidth]{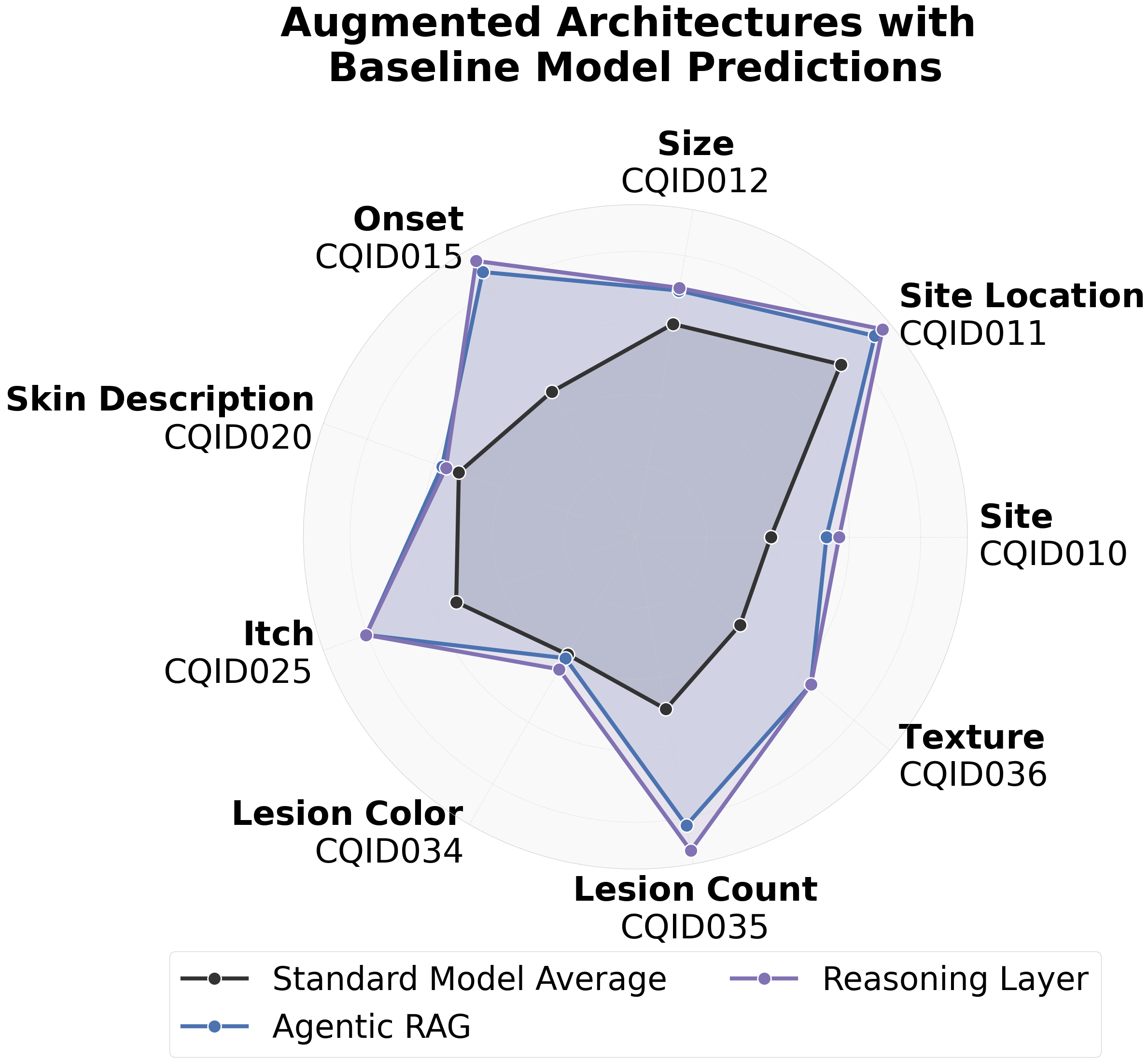}
    \includegraphics[width=0.42\textwidth]{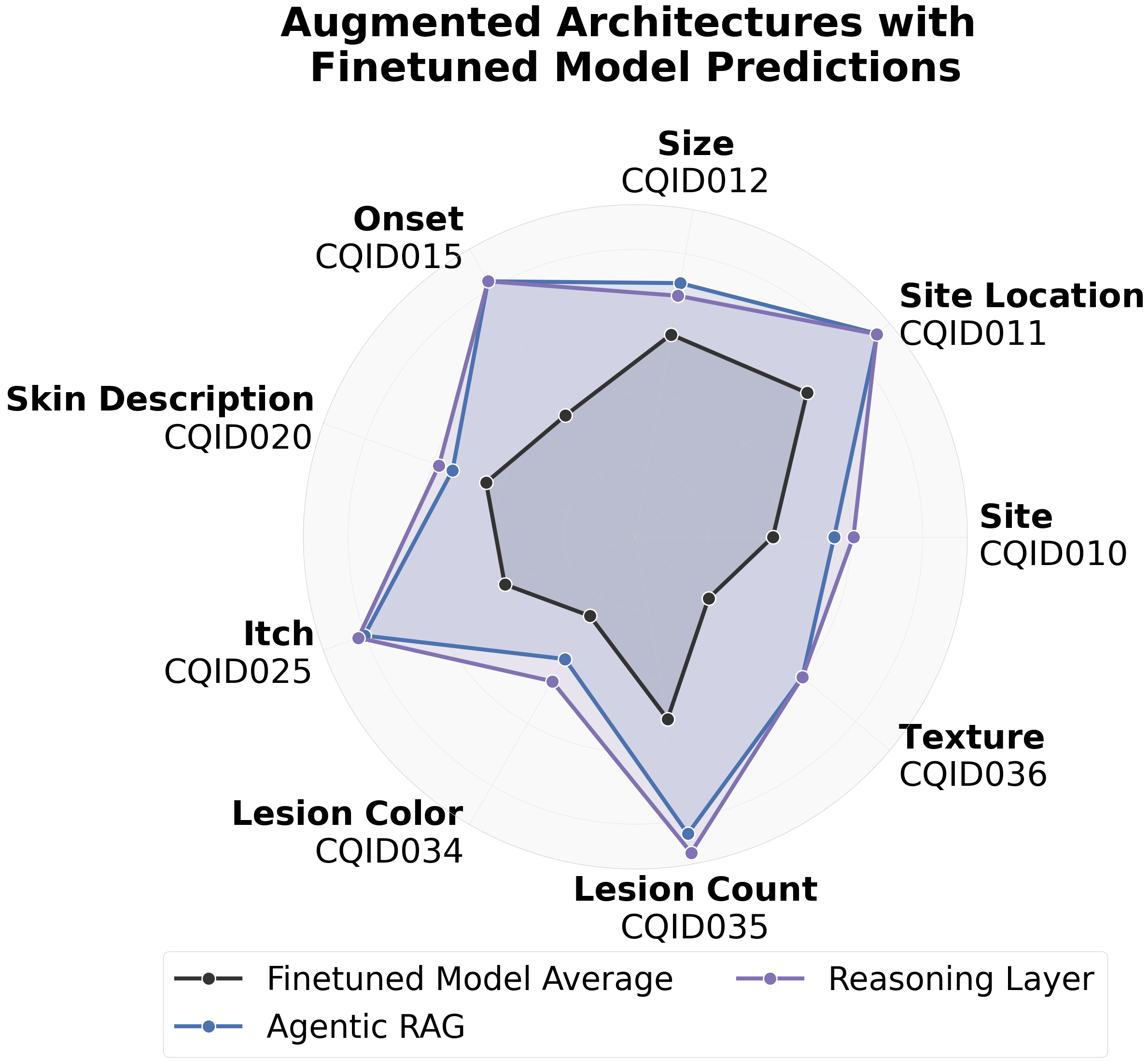}
    \caption{Average model performance versus architectural enhancements on validation data. Radar plots show mean accuracy across all baseline models (left) and all fine-tuned models (right) compared to Reasoning Layer and Agentic RAG, which aggregate predictions from these models. Both architectures consistently outperform model averages across question types.}
    \label{fig:special_model_comparison}
\end{figure}

\begin{figure}[H]
    \centering
    \includegraphics[width=0.42\textwidth]{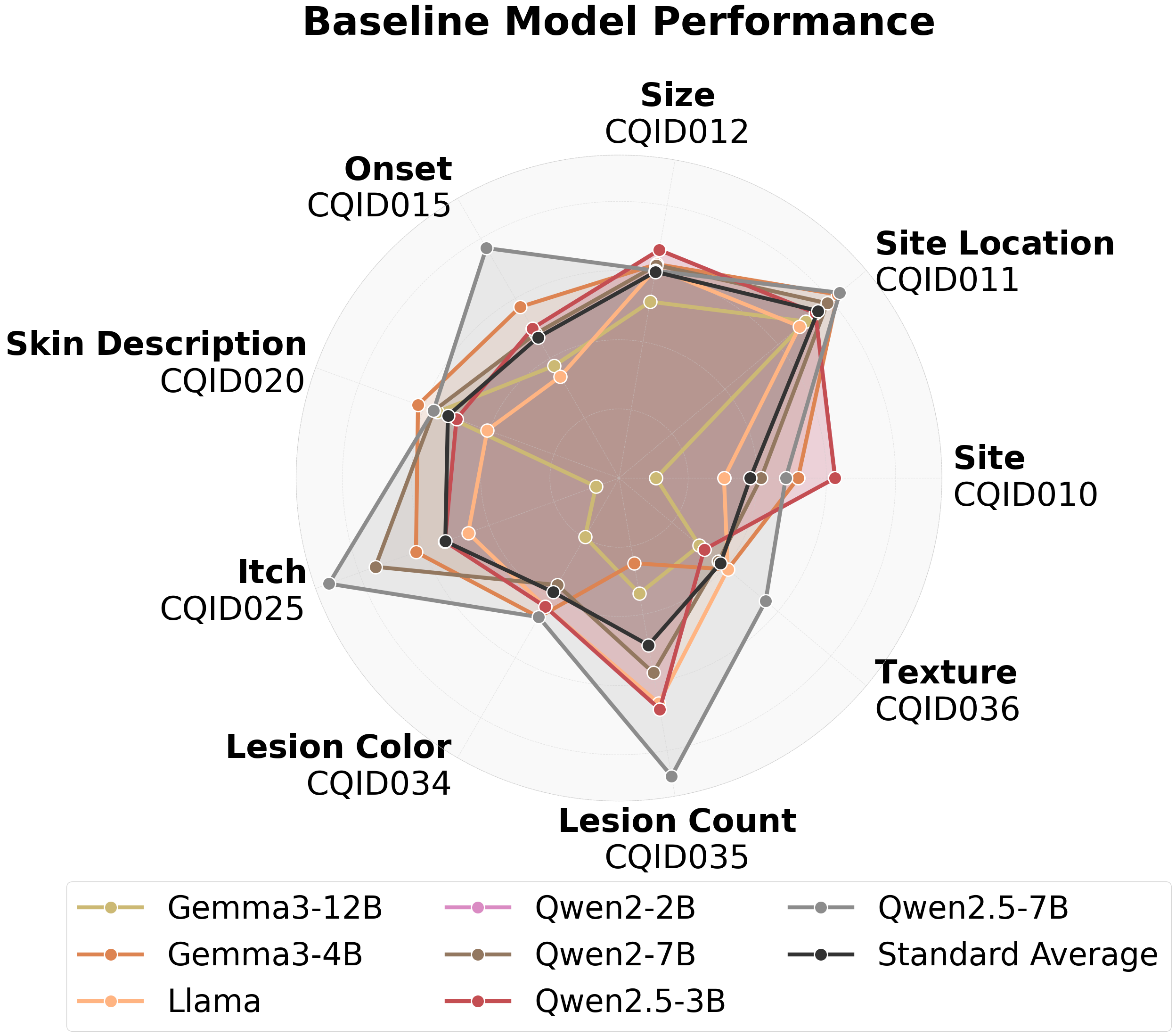}
    \includegraphics[width=0.42\textwidth]{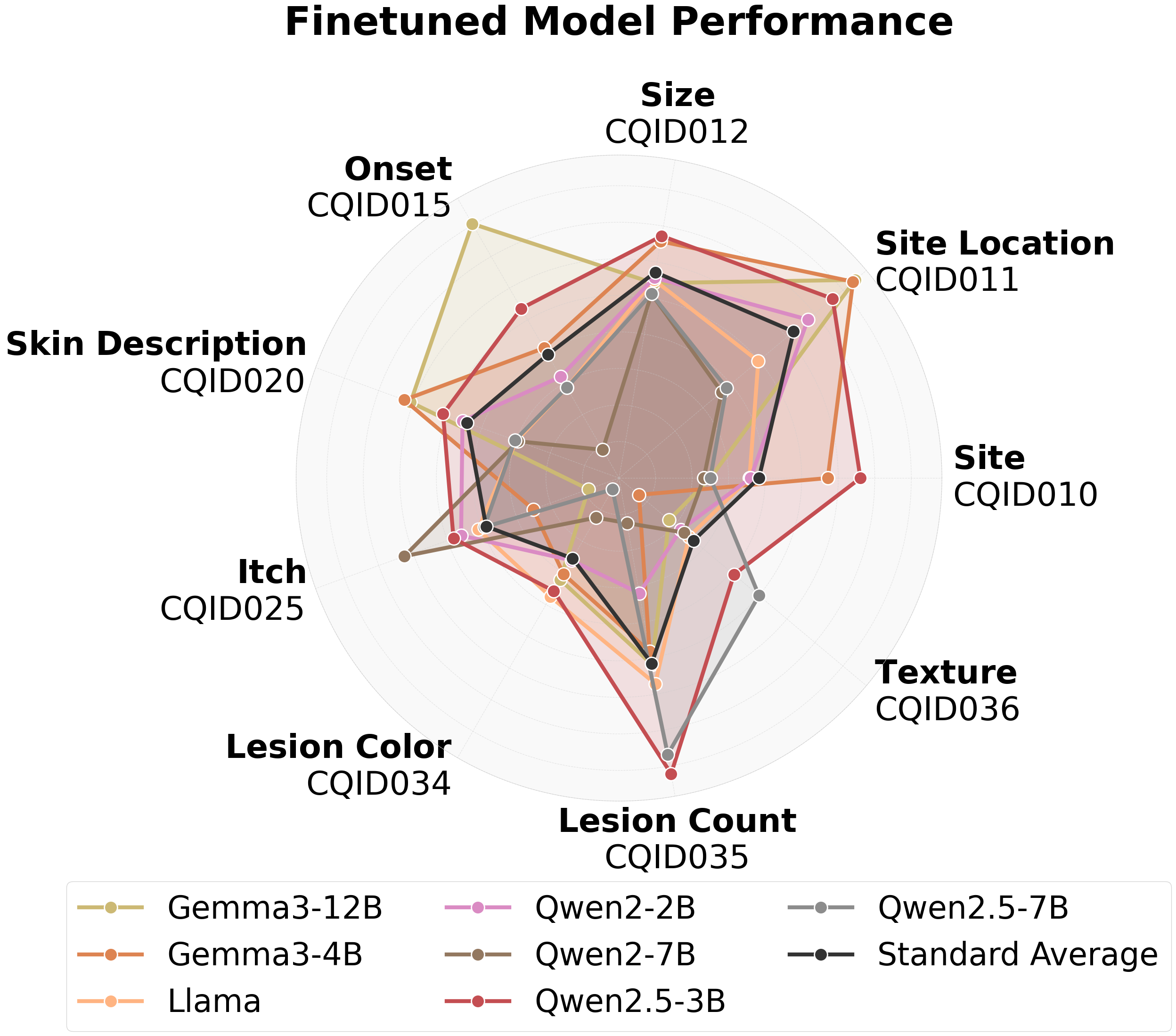}
    \caption{Individual model performance on validation data before (left) and after (right) fine-tuning. Radar plots reveal mixed effects: Gemma3-12B improves substantially while Qwen2.5-VL-7B degrades across most questions, demonstrating fine-tuning's unpredictable impact on different architectures.}
    \label{fig:base_vs_finetuned}
\end{figure}

\begin{figure}[H]
    \centering
    \includegraphics[width=0.8\linewidth]{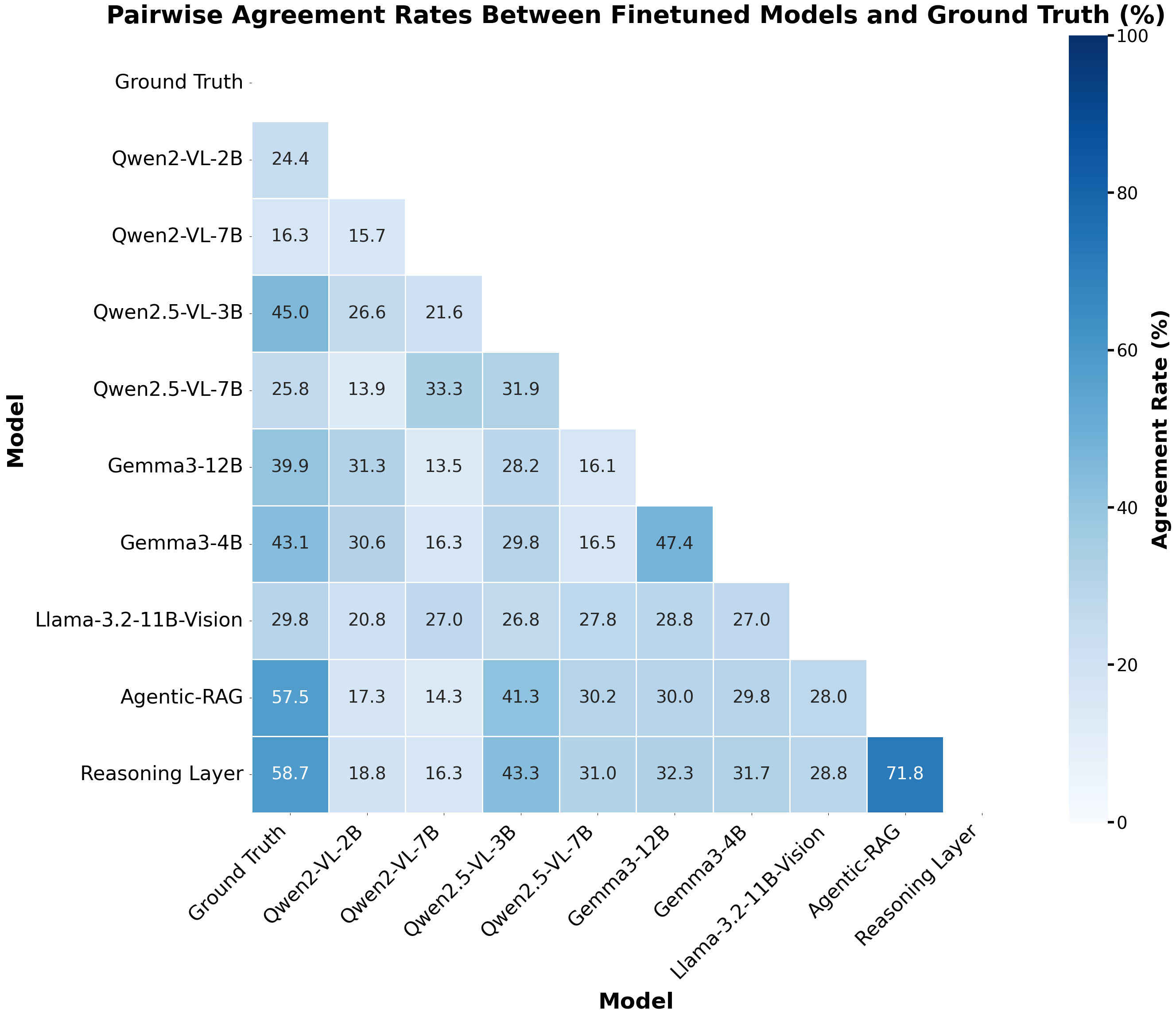}
    \caption{Pairwise agreement rates (\%) among finetuned model predictions, augmented architectures, and ground truth labels. Higher values indicate greater consistency in labels.}
    \label{fig:output2}
\end{figure}

\begin{table}[H]
\centering
\begin{tabular}{|p{3.5cm}|p{12.5cm}|}
\hline
\textbf{Field} & \textbf{Content} \\
\hline
\textbf{Query Context} & \textbf{Main Question:} What label best describes the affected area? \newline
\textbf{Type:} Skin Description \newline
\textbf{Category:} Skin Specific \newline
\textbf{Background:} Would brothers and sister here please help. These are my friend's hands. Has been to a few hospitals, but unable to get treated. Hope that people here can take a look to see what it is and what treatment is needed. Alternative medical treatment is also welcome. Thanks a lot. \newline
\textbf{Options:} raised or bumpy, flat, skin loss or sunken, thick or raised, thin or close to the surface, warty, crust, scab, weeping, Not mentioned \\
\hline
\textbf{Fine-tuned Model Predictions} & 
\textbf{Qwen2-VL-2B}: skin loss or sunken, warty, flat, \textbf{Qwen2-VL-7B}: raised or bumpy, thin or close to the surface, \textbf{Qwen2.5-VL-3B}: skin loss or sunken, warty, raised or bumpy, thick or raised, \textbf{Qwen2.5-VL-7B}: raised or bumpy, thick or raised, warty, \textbf{Gemma-3-4B}: crust, crust flat not mentioned, scab, thin or close to the surface, warty, \textbf{Gemma-3-12B}: thick or raised, crust, raised or bumpy, \textbf{LLaMA-3.2-11B}: thick or raised, crust, raised or bumpy, scab, bumpy, warty \\
\hline
\textbf{Reasoning} & 
The image analysis provides a detailed description of the affected areas across different sites (fingers, foot, ankle/lower leg). The 'SKIN\_DESCRIPTION' section states that the morphology varies by location. On the fingers, lesions are described as 'flat or slightly raised papules/confluent areas with dry skin and scaling'. This supports the options 'flat' and 'raised or bumpy'. On the foot, lesions are described as 'discrete, raised papules... some crusted and excoriated'. This supports 'raised or bumpy', 'crust', and 'scab'. On the ankle/lower leg, there are 'erythematous plaques with thick scaling, crusting, excoriations'. Plaques are elevated and often thickened, supporting 'raised or bumpy' and 'thick or raised'. 'Thick scaling' also supports 'thick or raised'. 'Crusting' supports 'crust' and 'scab'. Therefore, based on the evidence, the affected areas exhibit morphologies described by multiple labels from the options: flat, raised or bumpy, thick or raised, crust, and scab. These are all explicitly mentioned features of the affected skin across different sites. \\
\hline
\textbf{Final Answer} & \textbf{raised or bumpy, flat, thick or raised, crust, scab} \\
\hline
\end{tabular}
\caption{Reasoning Layer Output for Skin Description Task (CQID020) and Encounter 914.}
\label{tab:rl-ex}
\end{table}

\begin{table}[H]
\centering
\begin{tabular}{|p{3.5cm}|p{12.5cm}|}
\hline
\textbf{Field} & \textbf{Content} \\
\hline
\textbf{Query Context} & \textbf{Main Question:} What label best describes the affected area? \newline
\textbf{Type:} Skin Description \newline
\textbf{Category:} Skin Specific \newline
\textbf{Background:} Would brothers and sister here please help. These are my friend's hands. Has been to a few hospitals, but unable to get treated. Hope that people here can take a look to see what it is and what treatment is needed. Alternative medical treatment is also welcome. Thanks a lot. \newline
\textbf{Options:} raised or bumpy, flat, skin loss or sunken, thick or raised, thin or close to the surface, warty, crust, scab, weeping, Not mentioned \\
\hline
\textbf{Fine-tuned Model Predictions} & 
\textbf{Qwen2-VL-2B}: skin loss or sunken, warty, flat, \textbf{Qwen2-VL-7B}: raised or bumpy, thin or close to the surface, \textbf{Qwen2.5-VL-3B}: skin loss or sunken, warty, raised or bumpy, thick or raised, \textbf{Qwen2.5-VL-7B}: raised or bumpy, thick or raised, warty, \textbf{Gemma-3-4B}: crust, crust flat not mentioned, scab, , thin or close to the surface, warty, \textbf{Gemma-3-12B}: thick or raised, crust, raised or bumpy, \textbf{LLaMA-3.2-11B}: thick or raised, crust, raised or bumpy, scab, bumpy, warty \\
\hline
\textbf{Image and Clinical Context Analysis} & 
\textbf{Size}: Lesions range widely by location—1–3 mm papules on the foot and fingers, up to several cm patches on the lower leg and fingers. \newline
\textbf{Site Location}: Involves dorsal/lateral fingers, medial malleolus and anterior distal lower leg/ankle, dorsal feet/toes, and interdigital web spaces. Some fingers are unaffected. \newline
\textbf{Skin Description}: Erythematous, dry, and scaly fingers; crusted plaques with punctate bleeding and excoriations on the lower leg/ankle; reddish-brown papules/nodules with central crusting on the foot/toes. Possible vesicles on some fingers. \newline
\textbf{Lesion Color}: Varies by site—mild erythema with whitish scales on fingers, yellowish-white scales/crusts and puncta on lower leg, reddish-brown papules and darker crusting on feet/toes. \newline
\textbf{Lesion Count}: Extensive involvement—diffuse finger changes, 5–10+ distinct lesions on lower leg/ankle, 20+ papules/nodules on foot/toes, clusters on fingers. \newline
\textbf{Extent}: Localized to fingers, lower legs, and feet as shown. Some hands partially unaffected. Full body extent not assessed. \newline
\textbf{Texture}: Consistently rough or bumpy—due to scaling and crusts (fingers, lower leg) and papules/nodules with crusting or excoriation (feet/toes). \newline
\textbf{Onset Duration}: Appears subacute to chronic, with signs of flaring. History of prior visits suggests long-standing or recurrent nature. \newline
\textbf{Symptoms}: Visual signs of pruritus include excoriations and crusts. Seeking treatment implies symptomatic discomfort. \newline
\textbf{History}: Previous unsuccessful treatment attempts across multiple hospitals. \\
\hline
\textbf{Diagnosis Extractor} & \textbf{Diagnosis:} Lichen \newline
\textbf{Confidence:} 0.70 \newline
\textbf{Source:} Image analysis \\
\hline
\textbf{Retrieved Knowledge} & \textbf{Topic 1:} Lichen planus overview \newline
\textbf{Relevance Score:} 2.47 \newline
\textbf{Key Information:} \newline
-- Affects skin, mouth, genitals, scalp, and nails; may appear in one or several locations \newline
-- Can alter the appearance of fingernails and toenails \newline
-- Often develops on the wrists but may involve other areas of the body \newline
-- Not contagious and not a form of cancer \newline
\newline
\textbf{Topic 2:} Lichen planus diagnosis and treatment \newline
\textbf{Relevance Score:} 2.25 \newline
\textbf{Key Information:} \newline
-- Diagnosis may include physical exam, skin biopsy, and blood tests \newline
-- Treatment focuses on reducing symptoms and accelerating skin healing \newline
-- Topical and oral corticosteroids, antihistamines, PUVA therapy, and retinoids are options \newline
-- Mouth involvement may require different management approaches \newline
-- Skin cases usually resolve within two years; recurrence is possible \newline
-- Residual pigmentation changes may persist and require cosmetic treatments \newline
\newline
\textbf{Topic 3:} Neurodermatitis overview \newline
\textbf{Relevance Score:} 1.90 \newline
\textbf{Key Information:} \newline
-- Characterized by intense, localized itching often affecting arms, legs, scalp, or groin \newline
-- Itching may worsen during rest or nighttime and continue even without stress triggers \newline
-- Persistent scratching can lead to thickened skin, bleeding, and secondary infections \newline
-- Common signs include chronic itch, skin damage from scratching, and disturbed sleep \newline
-- Diagnosis is based on clinical signs and ruling out other causes of chronic itch \newline
\newline
\textbf{Topic 4:} Lichen planus signs and symptoms \newline
\textbf{Relevance Score:} 1.61 \newline
\textbf{Key Information:} \newline
-- Appears as shiny, firm, reddish-purple bumps often with white lines (Wickham's striae) \newline
-- May present as thick plaques, open sores, or scattered bumps on areas like wrists, back, or ankles \newline
-- Can affect nails (causing splitting or thinning) and the scalp (causing irritation or scarring hair loss) \newline
-- Oral form includes painful sores, redness, peeling gums, and white streaks or dots \newline
\newline
\textbf{Topic 5:} Neurodermatitis self-care strategies \newline
\textbf{Relevance Score:} 1.02 \newline
\textbf{Key Information:} \newline
-- Daily management includes cool compresses, colloidal oatmeal baths, and antihistamines \newline
-- Fragrance-free moisturizers and barrier protection reduce flare-ups \newline
-- Covering affected areas and keeping nails short can minimize scratching damage \newline
-- Heat, stress, and synthetic fabrics are common triggers to avoid \newline
-- Consistent follow-up with dermatology improves long-term outcomes \newline
-- Medical treatment is typically required for sustained symptom relief \\
\hline
\end{tabular}
\label{tab:agr-ex-3}
\end{table}

\begin{table}[H]
\centering
\begin{tabular}{|p{3.5cm}|p{12.5cm}|}
\hline
\textbf{Medical Knowledge Insights} & 
The retrieved medical knowledge discusses Lichen Planus and Neurodermatitis (Lichen Simplex Chronicus). Lichen Planus (Result 1, 4) can affect various sites including skin, nails, mouth, and has characteristic shiny, reddish-purple papules/plaques with white lines, often itchy. Neurodermatitis/LSC (Result 3, 5) is characterized by intense itching in one or two areas, leading to thickened, leathery skin from scratching, often on arms, legs, neck, scalp, groin. The visual findings of scaly/crusted patches/plaques on the lower leg (Image 2) and particularly the intensely pruritic-appearing, excoriated papules/nodules on the foot/toes (Image 3) are highly suggestive of Lichen Simplex Chronicus or a form of eczematous dermatitis driven by scratching. \newline
The diffuse scaling/dryness on other fingers (Image 1) could represent a different process or milder involvement. Image 5's potential vesicles could suggest dyshidrotic eczema. While classical Lichen Planus morphology (shiny, purple papules, Wickham's striae) is not prominent in these images, Lichen Planus can have varied presentations (hypertrophic LP, bullous LP) or co-exist with other conditions. The history of multiple failed treatments aligns with conditions like chronic eczema or LSC, which can be challenging to manage and often recur (Result 2, 3, 5). The intense itching visually evident is a key symptom of both LP and Neurodermatitis, prompting consideration of these diagnoses. Self-care measures for Neurodermatitis (Result 5) highlight the importance of breaking the itch-scratch cycle, which appears crucial based on the visual findings of excoriation.
\\
\hline
\textbf{Initial Reasoning} & 
The question asks for the best label(s) to describe the affected area based on the provided evidence. I need to review the `SKIN\_DESCRIPTION`, `TEXTURE`, and `SIZE` fields in the integrated evidence, as these directly describe the appearance of the lesions. The `SKIN\_DESCRIPTION` mentions:- Diffuse erythema, dryness, and scaling on fingers. Erythematous patches/plaques with yellowish-white scales, crusting, erosions, punctate bleeding, smaller papules, and excoriations on the lower leg/ankle.- Multiple discrete erythematous to reddish-brown papules/nodules, some with central crusting/excoriation on foot/toes. Small papules/vesicles on some fingers. The `TEXTURE` is described as \``Roughness... due to scaling/dryness..., scales/crusts/erosions..., and papules/nodules with crusted/excoriated surfaces... Some fingers... may have bumpy texture from papules/vesicles. The `SIZE` mentions \``papules/patches\" and \``papules/nodules\", indicating elevated lesions. \\
& Evaluating the options: `raised or bumpy`: Supported by the mention of papules, nodules, vesicles, plaques, and bumpy texture. `flat`: Patches are mentioned, which are flat, but the description also includes scaling, crusting, papules, nodules, and plaques, which are not flat. This isn't the best overall description of the *affected area* which contains multiple lesion types. `skin loss or sunken`: Supported by the mention of erosions and excoriations, which represent skin loss. However, this describes a specific feature within some lesions, not the overall description of the raised/thickened areas. `thick or raised`: Supported by the mention of plaques and the medical knowledge linking findings to conditions like Neurodermatitis which cause skin thickening. Papules and nodules are also raised. `thin or close to the surface`: Not supported by the evidence. Thickening is implied by plaques and the likely diagnosis. `warty`: Not mentioned or suggested by the descriptions. `crust`: Explicitly mentioned as present on the lower leg/ankle and foot/toes. `scab`: Implied by the presence of crusting, erosions, and punctate bleeding. A scab is essentially a crust over a wound. `weeping`: Not explicitly mentioned as a prominent feature. `Not mentioned`: Several features are mentioned. \\
& Based on the evidence, the affected areas exhibit a variety of morphologies. Key features include raised lesions (papules, nodules, plaques), surface changes (scaling, crusting), and consequences of scratching (erosions, excoriations, bleeding, likely scabs). Therefore, multiple labels are appropriate to best describe the diverse appearance. `raised or bumpy` is a good general term for the elevated lesions. `thick or raised` captures the larger, possibly thickened plaques and the potential for chronic thickening. `crust` is a direct description of a common surface finding. `scab` is a very likely finding given the excoriations and bleeding leading to crust formation. All four of these options are well-supported by the integrated evidence describing different aspects of the affected areas.\\
\hline
\textbf{Initial Results}
& \textbf{Predicted Answer:} raised or bumpy, thick or raised, crust, scab (Confidence: 0.74). \newline
\textbf{Uncertainty factors:}  \\
& -- Potential overlap or redundancy between 'raised or bumpy' and 'thick or raised'., \\ & -- Inferring the presence of 'scabs' from 'crusting, erosions, and punctate bleeding' rather than explicit mention. \\
& \textbf{Counterfactual:} If the evidence only described 'flat erythematous patches with scaling', then 'flat' would be a better descriptor and 'raised or bumpy', 'thick or raised', 'crust', and 'scab' would be less appropriate or incorrect. If the evidence explicitly stated the absence of crusting or excoriations, then 'crust' and 'scab' would not be selected. \\
& \textbf{Knowledge Contribution:} The medical knowledge about Neurodermatitis (Lichen Simplex Chronicus) reinforces the relevance of 'thick or raised' by explaining that intense scratching can lead to thickened skin (lichenification), which aligns with the description of plaques and the visually evident excoriations. This adds weight to 'thick or raised' as a key descriptor beyond just elevated lesions. \\
\hline
\end{tabular}
\label{tab:agr-ex-3}
\end{table}

\begin{table}[H]
\centering
\begin{tabular}{|p{3.5cm}|p{12.5cm}|}
\hline
\textbf{Reflection + Revision} & 
\textbf{Reflection:} The initial reasoning correctly identified the key evidence fields (`SKIN\_DESCRIPTION`, `TEXTURE`, `SIZE`) and extracted relevant descriptions of the lesions (scaling, crusting, erosions, papules, nodules, plaques, bumpy texture). It appropriately linked these findings to several of the options, correctly identifying support for `raised or bumpy`, `thick or raised`, and `crust`. The inference of `scab` from crusting, erosions, and bleeding is also a reasonable interpretation, as scabs are composed of dried blood and serum, often forming over erosions or excoriations. The reasoning correctly dismissed `flat` as not representative of the entire affected area, and correctly dismissed `thin or close to the surface`, `warty`, and `weeping` due to lack of evidence. However, the reasoning for excluding `skin loss or sunken` was weak. While it correctly noted that erosions and excoriations (representing skin loss) are features *within* some lesions, it failed to fully appreciate that these features are widespread and significant descriptors of the affected area, particularly on the lower leg and foot/toes. The question asks to describe the 'affected area', which includes all observed pathologies, not just the raised or thickened components. Skin loss due to scratching is a prominent characteristic visually evident in the provided images and described in the evidence, making 'skin loss or sunken' a valid and important descriptor of parts of the affected area. \\
& \textbf{Reasoning Gap:} The primary reasoning gap was the assumption that a descriptor must apply uniformly to the entire 'affected area' to be selected. The affected area encompasses diverse morphologies across different sites. Descriptors like 'skin loss or sunken', 'crust', 'scab', 'raised or bumpy', and 'thick or raised' all describe different, co-existing features within the overall affected region. The initial reasoning correctly applied this principle to select multiple 'raised' descriptors and 'crust', but inconsistently applied it by dismissing 'skin loss' because it didn't describe the *raised* areas. \\
& \textbf{Revised Answer:} raised or bumpy, skin loss or sunken, thick or raised, crust, scab. \\
& \textbf{Revised Confidence:} 0.90 \\
& \textbf{Revised Reasoning:} The question asks for the best label(s) to describe the affected area based on the integrated evidence. I need to perform a detailed analysis of the visual descriptions (`SKIN\_DESCRIPTION`, `TEXTURE`, `SIZE`) across the different affected sites mentioned (fingers, lower leg/ankle, foot/toes) and consider how these features relate to the provided options. The reflection highlighted the importance of considering the diversity of lesions within the 'affected area' and ensuring that features like skin loss from excoriation are not overlooked. The `SKIN\_DESCRIPTION` reveals a heterogeneous presentation across body sites: - Fingers: diffuse erythema, dryness, scaling, possibly small papules/vesicles. - Lower leg/ankle: erythematous patches/plaques with scales, crusting, erosions, punctate bleeding, smaller papules, excoriations. - Foot/toes: discrete papules/nodules, some with central crusting/excoriation. \\ & The `TEXTURE` is described as rough due to scaling, crusts, erosions, papules, and nodules, and potentially bumpy from papules/vesicles.The `SIZE` includes papules (1-3mm), nodules, patches, and plaques (up to several cm). Evaluating the options based on these descriptions: 1.  **raised or bumpy**: This is strongly supported. Papules, nodules, and plaques are explicitly mentioned, all of which are elevated lesions. The texture is described as rough and potentially bumpy from papules/vesicles. This applies to lesions on fingers (papules/vesicles), lower leg/ankle (papules, plaques), and foot/toes (papules/nodules). 2.  **flat**: Patches are mentioned on the lower leg/ankle. Patches are typically flat areas of altered color or texture. However, these patches have additional features like scales, crusting, erosions, and papules on them, meaning the surface is not uniformly flat. The diffuse erythema on fingers might be primarily flat, but is associated with scaling. Given the widespread presence of raised lesions (papules, nodules, plaques) and surface irregularities (scaling, crusting, erosions, excoriations), 'flat' is not the best overall descriptor for the *affected area* encompassing all these sites and morphologies. 3.  **skin loss or sunken**: This is strongly supported. 'Erosions' and 'excoriations' are explicitly mentioned on the lower leg/ankle and foot/toes. Erosions are superficial skin loss. Excoriations are deeper abrasions caused by scratching, also representing skin loss. The term 'skin loss' is directly applicable to these features. While the lesions aren't described as 'sunken' (which often implies atrophy or subcutaneous volume loss), the 'skin loss' part of the option is clearly present and significant, particularly in areas where scratching is severe (visually evident and supported by `SYMPTOMS`). 4.  **thick or raised**: This is strongly supported. Papules, nodules, and plaques are all 'raised'. Plaques on the lower leg/ankle suggest thickened skin. The medical knowledge about Neurodermatitis/Lichen Simplex Chronicus (`MEDICAL\_KNOWLEDGE\_INSIGHTS`) reinforces that chronic scratching leads to skin thickening (lichenification), which aligns with the description of plaques and the visual evidence of intense pruritus (excoriations). Thus, 'thick' is a highly relevant descriptor for some areas, particularly the lower legs, and 'raised' applies across all sites with papules/nodules/plaques. 5.  **thin or close to the surface**: 'Thin' is not supported by any evidence; in fact, thickening is suggested. Some lesions like erosions or superficial papules are 'close to the surface', but 'thin' as a descriptor for the skin texture is inaccurate. 6.  **warty**: There is no description of warty appearance (i.e., papillomatous, verrucous surface). 7.  **crust**: This is explicitly mentioned as a feature on the lower leg/ankle and foot/toes. This is a direct and accurate descriptor of a surface characteristic in the affected area. 8.  **scab**: This is strongly implied. The presence of 'crusting', 'erosions', and 'punctate bleeding' indicates areas where the skin surface has been broken (erosions, excoriations) and dried blood, serum, and exudate have formed a crust. \\ & A scab is precisely such a formation over a wound or abrasion. This is a very likely finding given the described features. 9.  **weeping**: Not explicitly mentioned as a prominent feature. While some serous exudate leading to crusting might occur, 'weeping' typically implies active discharge, which is not described. Multiple options are necessary to accurately describe the varied affected areas. The analysis supports `raised or bumpy`, `skin loss or sunken`, `thick or raised`, `crust`, and `scab` as valid descriptors of different aspects of the observed lesions and affected regions. \\
& \textbf{Knowledge utilization assessment:} Medical knowledge about conditions like Lichen Simplex Chronicus (Neurodermatitis) was integrated to understand the significance of visually evident features like intense excoriation and described findings like plaques. This knowledge reinforces that chronic scratching leads to skin thickening (lichenification), validating 'thick or raised' as a descriptor, and also explains why skin damage (erosions, excoriations, leading to skin loss, crusts, and scabs) is a prominent feature in such pruritic conditions. This contextual understanding strengthens the evidence supporting descriptors like 'thick or raised', 'skin loss or sunken', 'crust', and 'scab' in the context of the patient's history and visual presentation.\\
\hline
\textbf{Final Answer} & \textbf{raised or bumpy, skin loss or sunken, thick or raised, crust, scab} \hfill \textbf{Confidence:} 0.90 \\
\hline
\end{tabular}
\caption{Agentic RAG for Skin Description Task (CQID020) and Encounter 914.}
\label{tab:agr-ex-3}
\end{table}

\end{document}